\documentclass[runningheads]{llncs}
\usepackage{graphicx}
\usepackage{latexsym}
\usepackage{amsmath}
\usepackage{adjustbox}
\usepackage{caption}
\usepackage{multirow}
\usepackage{subcaption}
\usepackage{booktabs}
\usepackage{color}
\usepackage{amssymb}
\usepackage{xcolor}
\usepackage{hyperref}

\begin{document}
%
\title{SODAWideNet - Salient Object Detection with an Attention augmented Wide Encoder Decoder network without ImageNet pre-training}
\titlerunning{SODAWideNet}
%
\author{Rohit Venkata Sai Dulam\thanks{Corresponding Author} \and
Chandra Kambhamettu
}
\authorrunning{RVS Dulam\thanks{Corresponding Author} \and C Kambhamettu}
%
\institute{
Video/Image Modeling and Synthesis (VIMS) Lab\\
University of Delaware, Newark DE 19716, USA \\
\email{\{rdulam,chandrak\}@udel.edu}
}

\maketitle              
\begin{abstract}
Developing a new Salient Object Detection (SOD) model involves selecting an ImageNet pre-trained backbone and creating novel feature refinement modules to use backbone features. However, adding new components to a pre-trained backbone needs retraining the whole network on the ImageNet dataset, which requires significant time. Hence, we explore developing a neural network from scratch directly trained on SOD without ImageNet pre-training. Such a formulation offers full autonomy to design task-specific components. To that end, we propose SODAWideNet, an encoder-decoder-style network for Salient Object Detection. We deviate from the commonly practiced paradigm of narrow and deep convolutional models to a wide and shallow architecture, resulting in a parameter-efficient deep neural network. To achieve a shallower network, we increase the receptive field from the beginning of the network using a combination of dilated convolutions and self-attention. Therefore, we propose Multi Receptive Field Feature Aggregation Module (MRFFAM) that efficiently obtains discriminative features from farther regions at higher resolutions using dilated convolutions. Next, we propose Multi-Scale Attention (MSA), which creates a feature pyramid and efficiently computes attention across multiple resolutions to extract global features from larger feature maps. Finally, we propose two variants, SODAWideNet-S (3.03M) and SODAWideNet (9.03M), that achieve competitive performance against state-of-the-art models on five datasets. We provide the code and pre-computed saliency maps \href{https://github.com/VimsLab/SODAWideNet}{here}.

\end{abstract}
\section{Introduction}
2D Salient Object Detection (SOD) is a dense prediction task to identify objects of interest in images that attract humans' immediate attention. Earlier works on SOD used hand-crafted priors, while recently, the focus has shifted to learning-based approaches using Convolutional Neural Networks (CNN)~\cite{gao2020sod100kcsf,wu2022edn,Qin_2020_PRU2} and Transformers~\cite{jing_ebm_sod21generative,Liu_2021_ICCV_VST,xie2022pyramid}. 


SOD has vastly benefitted from multi-scale features extracted by pre-trained backbones~\cite{he2016deepresnet,simonyan2014veryvgg16,liu2021swin}, so most current works~\cite{wei2020f3net,Ke_2022_WACV_RCSB,liu2018picanet,qin2019basnet,yang2021progressivepsg,wu2022edn,wang2021pyramidtransf,jing_ebm_sod21generative,Liu_2021_ICCV_VST,gao2020sod100kcsf} build on top of them. Nevertheless, they have some drawbacks. Firstly, pre-trained backbones designed for image classification are trained on the ImageNet dataset. \cite{hermann2020origins} suggests that models trained on the ImageNet dataset utilize local information like textures and contrast information to classify objects. Since SOD requires a sound understanding of local and global features, ImageNet pre-training might be sub-optimal. Additionally, designing a deep learning model for Image Classification and developing new feature refinement modules to fine-tune a downstream task like SOD takes significant time and effort. Also, architecturally, the most famous pre-trained convolutional backbone for SOD, ResNet-50, uses a stack of small convolution kernels with identical receptive fields at each layer. Hence, attaining a global receptive field requires significant downsampling of the input, causing a dilution of essential features and an increase in parameters. Owing to these shortcomings, we propose a deep learning model for SOD that does not use ImageNet pre-training and performs comparatively against other state-of-the-art methods at a fraction of the parameters.  

\begin{figure*}[ht!]
  \includegraphics[width=\linewidth, keepaspectratio]{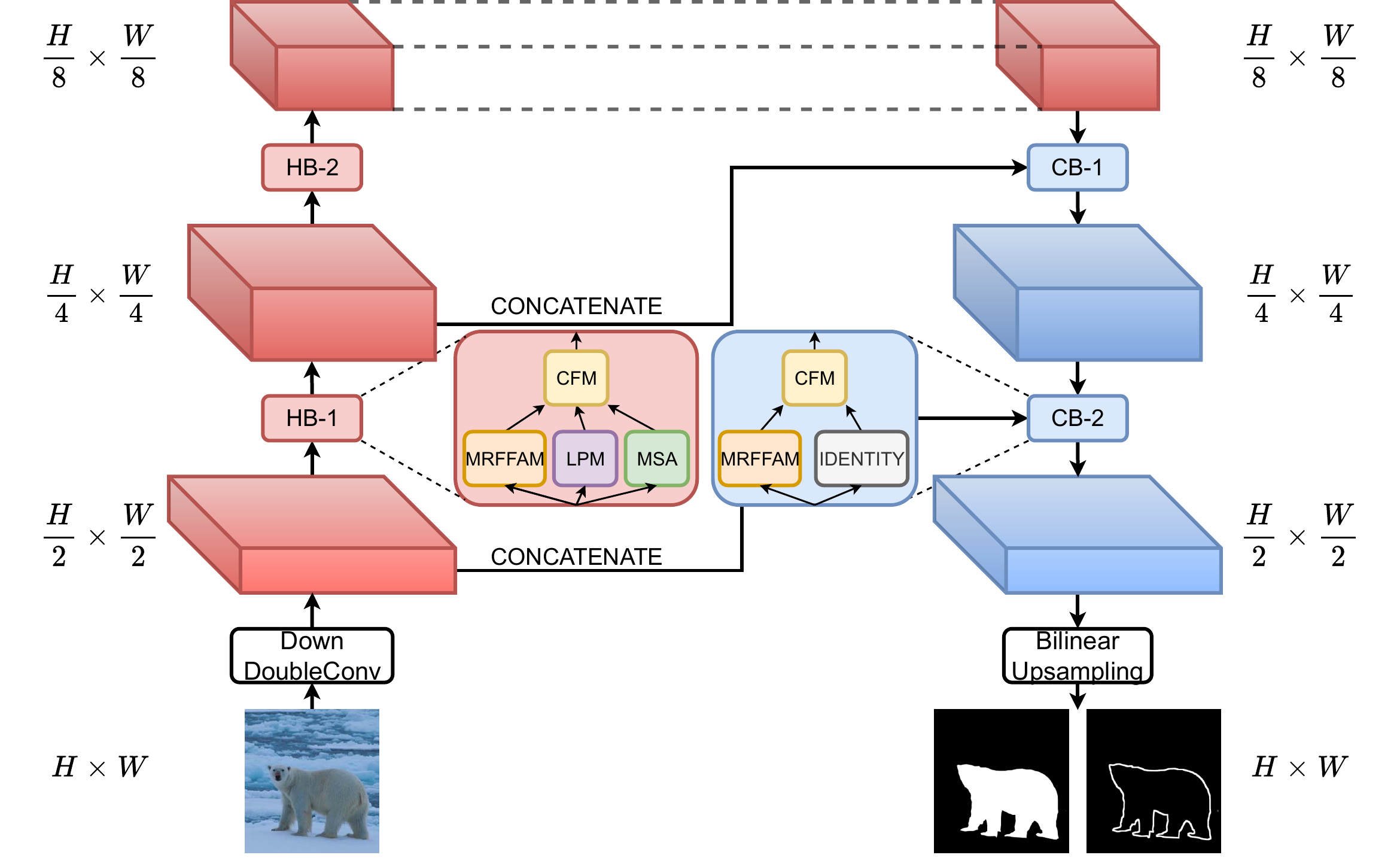}
  \caption{SODAWideNet. Each of the hybrid blocks (HB) consists of three parallel streams, namely MRFFAM, LPM, and MSA. Features extracted by these individual modules are input to the CFM, which becomes the output of the hybrid block. In the convolution block (CB), features are refined by the MRFFAM before upsampling. [Best viewed in color]}
  \label{fig:ModelArchitecture}
\vspace{-5mm}
\end{figure*}


Inspired by vision transformers' ability to attain a global receptive field at every layer, we propose a novel encoder-decoder-style neural network called SODAWideNet, which uses large convolutional kernels and self-attention for Salient Object Detection. Furthermore, to expand the receptive field at every layer of our network, we propose Multi-Receptive Field Feature Aggregation Module (MRFFAM), a fully convolutional module made of dilated convolutions to encode long-range dependencies. To increase the receptive field further, we employ self-attention in our model. Although very powerful, calculating attention is computationally intensive, especially for high-resolution feature maps. To remedy this shortcoming, we propose Multi-Scale Attention (MSA), which creates a feature pyramid using average pooling and then computes attention across scales. Local information, including contrast and texture, is also necessary to identify an object. Hence, we propose a Local Processing Module (LPM) to extract features from a local area using $3 \times 3$ convolutions. Finally, we use contour information as an auxiliary learning task to generate better saliency predictions. This use of contour detection changes the problem into multi-task learning, which helps our model learn more discriminative features beneficial to both SOD and contour detection.  
We briefly summarize our contributions below - 
\begin{enumerate}
    \item We propose \textbf{SODAWideNet} and \textbf{SODAWideNet-S}, two deep learning models that use large convolutional kernels and attention at every layer to extract long-range features without significant downsampling of the input.
    \item To efficiently extract and combine features from multiple receptive fields using dilated convolutions from larger resolutions, we propose Multi Receptive Field Feature Aggregation Module (MRFFAM).
    \item To compute self-attention on high-resolution inputs efficiently, we propose Multi-Scale Attention (MSA).
\end{enumerate}

\begin{figure}
    \centering
  \includegraphics[width=0.8\linewidth]{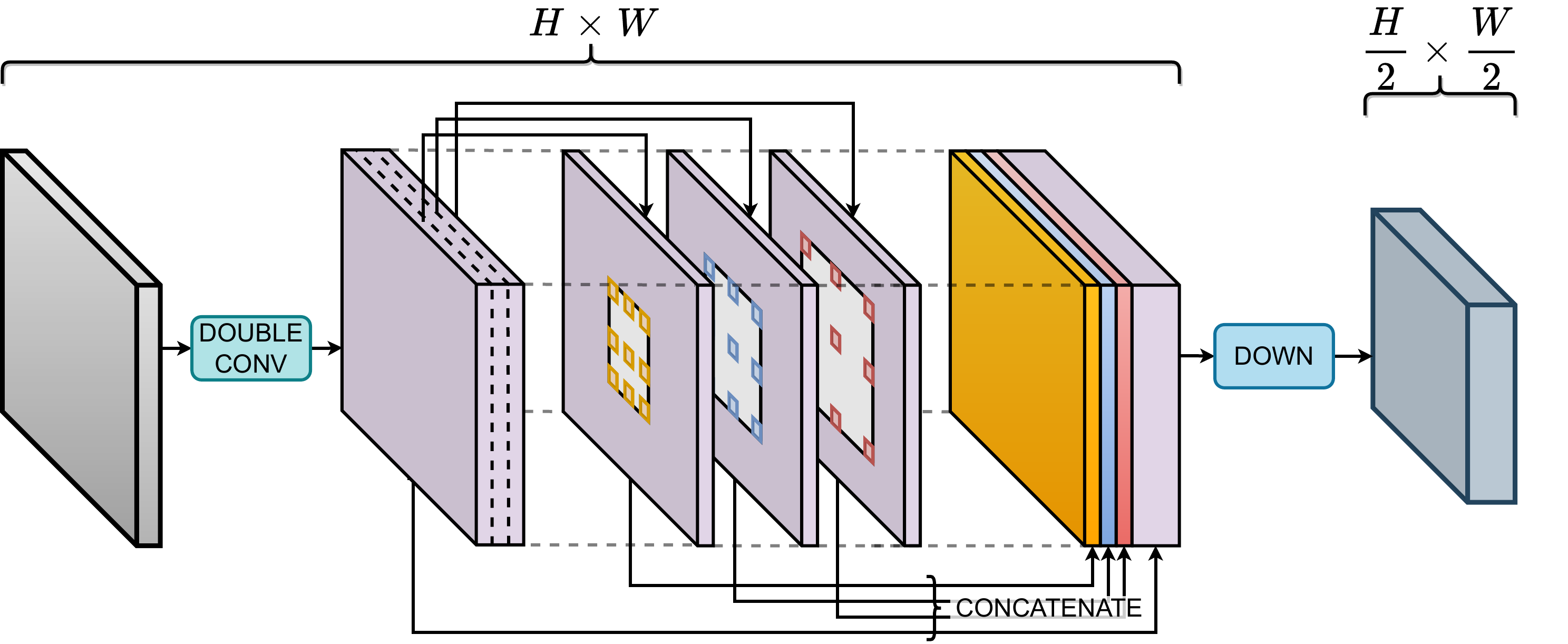}
  \caption{Multi Receptive Field Feature Aggregation Module (MRFFAM) [Best viewed in color]}
  \label{fig:MRFFAM}
\vspace{-12mm}
\end{figure}

\begin{figure}[t!]
\centering
  \includegraphics[width=\linewidth]{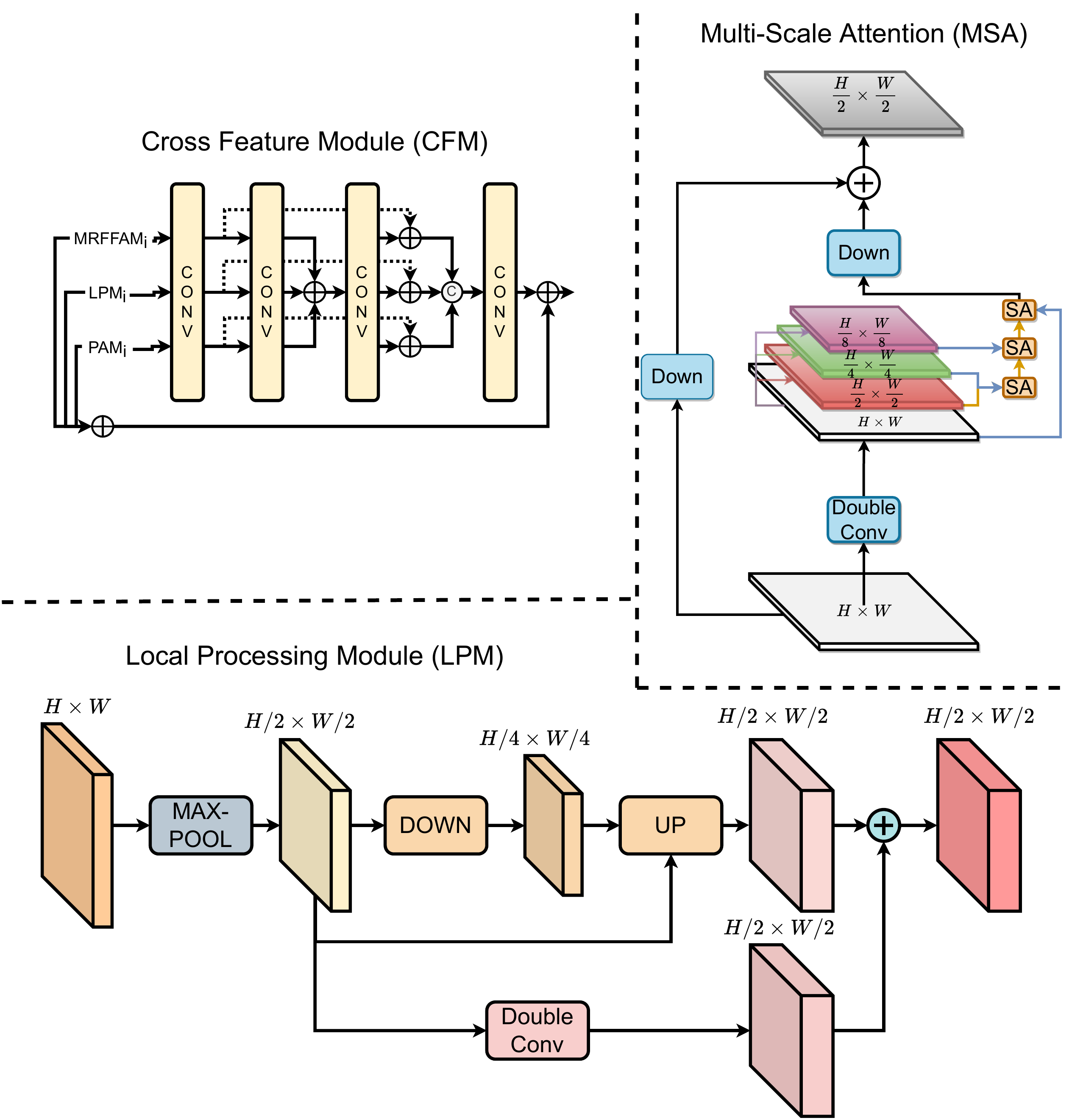}
  \caption{Cross Feature Module (CFM), Multi-Scale Attention (MSA), and Local Processing Module (LPM). In MSA, \textcolor[HTML]{D79B00}{$\longrightarrow$} indicates the feature map used to compute the keys and values, and\textcolor[HTML]{6C8EBF}{$\longrightarrow$} indicates the feature map to compute the queries. [Best viewed in color]}
  \label{fig:MSA}
\vspace{-5mm}
\end{figure}

\section{Related Works}

The first works in SOD used non-learning-based techniques whose performance was limited. Then came deep learning-based approaches based on fully convolutional neural networks that gave promising results.

Liu \textit{et al}.~\cite{liu2018picanet} (PiCANet-R) proposes a pixel-wise contextual attention module to selectively attend to informative context locations for each pixel from the features from the Resnet-50 backbone~\cite{he2016deepresnet}. Qin \textit{et al}.~\cite{qin2019basnet} (BASNet) uses a pre-trained ResNet34~\cite{he2016deepresnet} deep learning model and a boundary refinement module to generate saliency predictions. Wei \textit{et al}.~\cite{wei2020f3net} (F3-Net) use a cross-feature module (CFM) and cascaded feedback decoder (CFD) to generate saliency predictions using a pre-trained ResNet-50 model~\cite{he2016deepresnet}. Liu and Zhang \textit{et al}.~\cite{Liu_2021_ICCV_VST} (VST) propose a transformer-based SOD model using a T2T-ViT~\cite{Yuan_2021_ICCVT2T} pre-trained transformer. Yang \textit{et al}.~\cite{yang2021progressivepsg} use a progressive self-guided loss function that simulates a morphological closing operation on the model predictions to progressively create auxiliary training supervisions to guide the training process incrementally. They use the ResNet-50 model~\cite{he2016deepresnet} as a pre-trained backbone. Zhang \textit{et al}.~\cite{jing_ebm_sod21generative} (Generative Transformer) propose a vision transformer following an energy-based prior for salient object detection. Ke \textit{et al}.~\cite{Ke_2022_WACV_RCSB} (RCSB) uses contour information and a pre-trained ResNet-50 model~\cite{he2016deepresnet} to generate crisp boundaries for saliency prediction. Cheng \textit{et al}.~\cite{gao2020sod100kcsf} (CSF-Net) add cross stage fusion (CSF) to a Res2Net50~\cite{gao2019res2net} pre-trained on the ImageNet dataset~\cite{deng2009imagenet} to produce saliency predictions. Wu and Liu \textit{et al}.~\cite{wu2022edn} (EDNet) use an extreme downsampling technique to obtain high-level features essential for accurate saliency prediction. First, they pre-train their backbone on the ImageNet dataset~\cite{deng2009imagenet} before fine-tuning it for SOD. Xie \textit{et al}.~\cite{xie2022pyramid} (PGNet) generates saliency predictions by combining features from a pre-trained Resnet18 backbone~\cite{he2016deepresnet} and a Swin-B 224 transformer~\cite{liu2021swin} encoder to produce saliency predictions. Zhuge and Fan \textit{et al}.~\cite{zhuge2021salienticon} use a Swin-B-22k~\cite{liu2021swin} encoder to extract semantic features refined by novel feature aggregation modules. Lee \textit{et al}.~\cite{lee2022tracer} uses an EfficientNet~\cite{tan2019efficientnet} backbone to produce saliency predictions.
\section{Method}

Our model SODAWideNet, is shown in Figure~\ref{fig:ModelArchitecture} builds upon the famous U-Net~\cite{ronneberger2015UNET} deep learning architecture. In this section, we briefly introduce the U-Net architecture and list some of its components. We then go into details of individual pieces of the SODAWideNet model, namely MRFFAM, LPM, MSA, Hybrid, and Convolutional blocks. Finally, we describe the loss function used to train the proposed model. 

\subsection{Overview of U-Net}
U-Net is an encoder-decoder-style model that consists of a series of downsampling and upsampling layers. Below, we describe the downsampling block -

\begin{equation*}
    ConvB(x, d) = ReLU(BN(F_{3 \times 3, d}(x)))
\end{equation*}
\begin{equation*}
    Double\_Conv(x) = ConvB(ConvB(x, 1), 1)
\end{equation*}
\begin{equation*}
    Down(x) = Double\_Conv(max\_pool(x))
\end{equation*}

Similarly, the upsampling block can be described as shown below -
\begin{equation*}
    Up(x, y) = Double\_Conv(cat(Upsampling(x), y))
\end{equation*}
where $F_{3 \times 3, d}$, BN, ReLU, max\_pool, cat, and Upsampling imply a $3 \times 3$ convolution with dilation rate 'd', Batch Normalization, Rectified Linear Unit, Max-pooling, concatenation operation, and bilinear upsampling by a factor of two, respectively.


\subsection{Multi-Receptive Field Feature Aggregation Module (MRFFAM)}

Multi-Receptive Field Feature Aggregation module extracts and aggregates semantic information from multiple receptive fields. Similar to transformers which attain a global receptive field through self-attention, these convolution kernels obtain information from larger contexts. 
As shown in Figure~\ref{fig:MRFFAM}, at each layer, the input to MRFFAM is divided in the channel dimension and is input to various dilated convolutions with different dilation rates. Like \textit{Double\_Conv}, we use two dilated convolutions in series for each dilation rate. Dilation rates used at each layer are shown in Table~\ref{tab:DilationRates}, obtained through thorough experimentation.
\begin{table}[h!]
    \centering
\begin{tabular}{ |c|c|c|c|c| } 
\hline
Input Resolution & Layer & Dilation Rates & Output Resolution \\
\hline
$192 \times 192 \times 64 $ & HB1 & 6, 10, 14, 18, 22 & $96 \times 96 \times 128 $ \\
$96 \times 96 \times 128 $ & HB2 & 6, 10, 14, 18 & $48\times 48 \times 128 $ \\
\hline
$48 \times 48 \times 64 $ & CB2 & 6, 10, 14, 18 & $48 \times 48 \times 64 $ \\
$96 \times 96 \times 64 $ & CB3 & 6, 10, 14, 18, 22 & $96 \times 96 \times 64 $ \\
\hline
\end{tabular}
\caption{Configuration of MRFFAMs used at each block of SODAWideNet.}
\label{tab:DilationRates}
\vspace{-10mm}
\end{table}




\textbf{Distinction from ASPP module} - The primary difference between MRFFAM and ASPP is the location of these modules. ASPP extracts long-range features from the output of a pre-trained backbone, whereas MRFFAM is employed at each layer of our network, making it an essential component of the backbone. Secondly, dilated convolutions in MRFFAM work on a subset of the input, unlike ASPP, where every convolution operation processes the entire feature maps. Finally, our formulation provides scope for future works to explore different dilation rates for various input resolutions.



\subsection{Local Processing Module}
Local features like texture and contrast are essential to differentiate between foreground and background. Instead of solely relying on a single scale to extract local features, we use two different scales, as seen in Figure~\ref{fig:MSA}. Context from multiple scales enables the network to obtain richer local features. The successive max-pooling layers help obtain the most discriminative features from a smaller neighborhood which are further refined by the network.

\subsection{Multi Scale Attention (MSA)}

Self-attention is one of the most significant contributors to the success of vision transformers. This is because self-attention enables a global receptive field and introduces input dependency in the network. Unlike convolutional weights, which are frozen during inference, the dot product between queries and keys instills reliance on the input, forcing the network to extract semantically rich features. Although very powerful, calculating attention is computationally expensive. Hence, we create a feature pyramid and compute attention across multiple resolutions. To construct the feature pyramid, we reduce the spatial resolution of the input using average pooling and refine them using \textit{Double\_Conv}. Once created, we compute attention among the top two resolutions in the stack and continue the process until we reach the lowest resolution in the feature map stack. Finally, we calculate attention between the lowest resolution and the input of the feature map stack. The lowest resolution in each of the pyramids is $R^{H/16 \times W/16}$ where $H \times W$ is the spatial resolution of the input image. Figure~\ref{fig:MSA} illustrates the MSA module.
The keys and values are computed from higher resolution feature map, whereas the queries are computed from smaller resolution. Hence, the output's spatial resolution is the same as the query's.

\textbf{Differences from Spatial Reduction Attention (SRA)} - SRA uses strided convolutions to reduce the spatial resolution of feature maps before computing attention. However, strided convolutions can only summarize features effectively with large amounts of training data. Additionally, SRA reduces the resolution once and calculates attention between the queries and the downsampled keys. In contrast, we adopt a hierarchical approach to reducing the spatial resolution to compute attention, thus retaining essential features.
 
\subsection{Cross Feature Module (CFM)}

Features from the MRFFAM, LPM, and MSA entail varying semantic contexts. We modify
the CFM layer in \cite{wei2020f3net} to effectively combine them. The architecture is illustrated in Figure~\ref{fig:MSA}. The output at each layer of our model is the output of the CFM block. As seen from the architecture, each input passes through a series of \textit{Conv} layers which is similar to \textit{ConvB} but uses GroupNorm~\cite{wu2018group}. 

\subsection{Hybrid and Convolution blocks}

Although our model is an encoder-decoder-style network, the encoder is heavier than the decoder. Furthermore, each encoder layer is a hybrid block since it contains convolutional and attention modules. MRFFAM$_{i}$, LPM$_{i}$, and MSA$_{i}$ are the outputs of the previously proposed models. The output of a hybrid block is -

\begin{equation*}
    \text{HB}_{i} = \text{CFM}_{i}(\text{MRFFAM}_{i}, \text{LPM}_{i}, \text{MSA}_{i})
\end{equation*}

Similarly, each decoding block is called a convolution block due to only using the MRFFAM, which is a fully convolutional module. The only other operation in this block is the identity operation which indicates using the input \textit{X}$_{i}$ to the convolutional block as is. 

\begin{equation*}
    \text{CB}_{i} = \text{CFM}_{i}(\text{MRFFAM}_{i}, \text{X}_{i})
\end{equation*}





\subsection{Loss Function}

\begin{figure}[ht!]
    \centering
    \begin{subfigure}{0.2\linewidth}
    \includegraphics[width=1.0\linewidth]{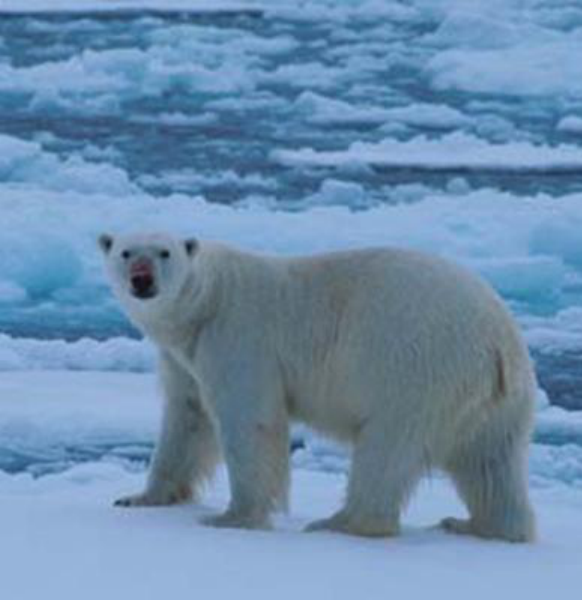}
  \end{subfigure}
  \begin{subfigure}{0.2\linewidth}
    \includegraphics[width=1.0\linewidth]{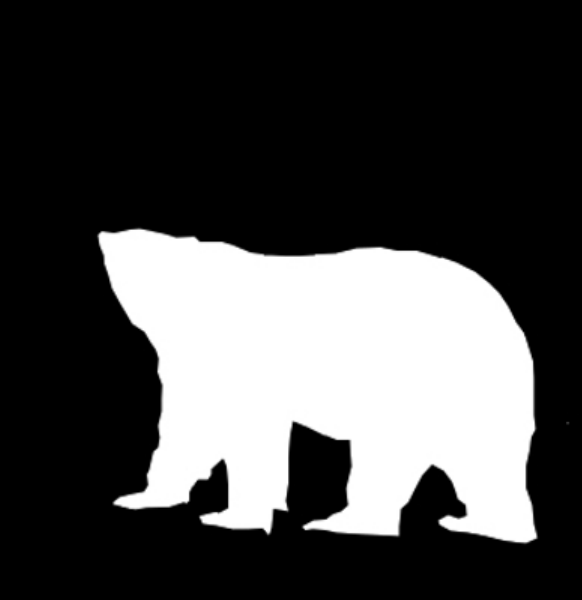}
  \end{subfigure}
  \begin{subfigure}{0.2\linewidth}
    \includegraphics[width=1.0\linewidth]{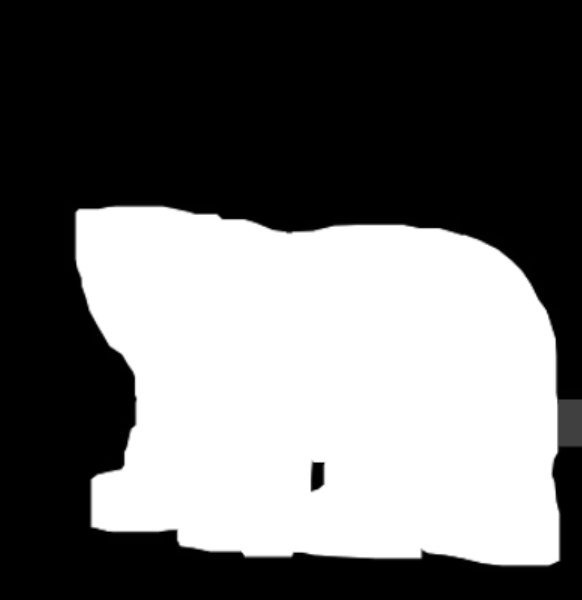}
  \end{subfigure}
  \begin{subfigure}{0.2\linewidth}
    \includegraphics[width=1.0\linewidth]{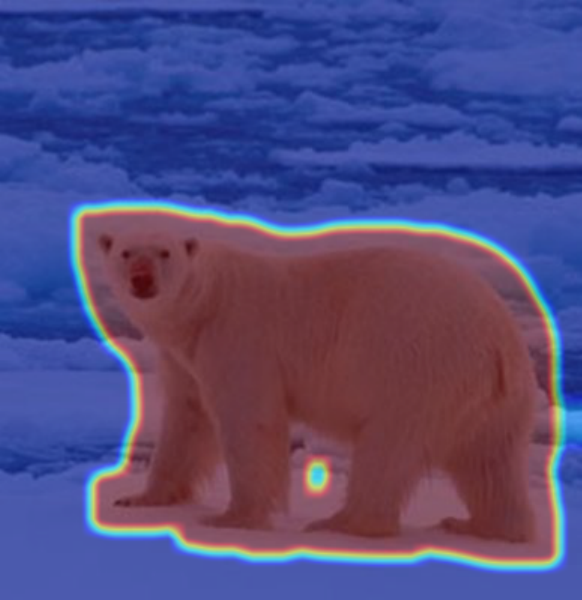}
  \end{subfigure}
    \caption{Given the image and its groundtruth in images one and two, $\alpha$ in equation~\ref{Equation_10} determines the weight for individual pixels, shown in three. Red indicates a more significant weight, whereas blue indicates a lower weight.}
    \label{fig:alpha}
\vspace{-7mm}
\end{figure} 

We modify the loss function proposed by \cite{wei2020f3net} which is a custom Weighted Binary Cross-Entropy loss (BCE) and IoU loss. The authors calculate a parameter $\alpha_{i,j}$ to assign weights for each pixel ($i,j$). It is the difference between the average values of all pixels in a window centered at a particular pixel and the center pixel's value. Instead, we use the maximum value in the window. Figure~\ref{fig:alpha} shows the weights assigned to each pixel. The third image shows each pixel's $\alpha_{i,j}$ value. Intuitively, the pixels of the salient object and its surroundings should have a higher weight, which we obtain through our formulation of $\alpha_{i,j}$. Equation \ref{Equation_10} shows the calculation of $\alpha$.

\begin{equation}\label{Equation_10}
    \alpha_{ij} = max(A_{ij})
\end{equation} where $A_{ij}$ represents the area surrounding pixel ($i,j$). Hence, the final loss function is defined as -
\begin{equation}\label{Equation_15}
    L_{Salient} = L_{wBCE} + L_{wiou} + L_{wL1} + L_{SSIM}
\end{equation}
where $L_{wiou}$ is the weighted IOU loss used in \cite{wei2020f3net}, $L_{wL1}$ is the L1-loss, and $L_{SSIM}$ is the SSIM loss. $`w'$ indicates that the loss value is calculated per pixel and then multiplied with that pixel's $\alpha_{i,j}$. Similarly, we use a weighted combination of BCE, Dice Loss~\cite{deng2018learningdice}, and SSIM loss for contour generation. 

\begin{equation}\label{Equation_18}
    L_{Contour} = 0.001 \cdot L_{bce} + L_{dice} + L_{SSIM}
\end{equation}

\section{Results and Ablation Experiments}

\begin{table*}[ht!]
  \caption{Comparison of our method with 15 other methods in terms of max F-measure $F_{max}$, MAE, and $E_{m}$ measures. The greater the value in all measures except MAE, the better performance. $^{*}$ indicates models trained from scratch on SOD instead of ImageNet pre-trained weights.}
  \label{tab:ResultsTable}
  \begin{adjustbox}{width=\linewidth}
  \begin{tabular}{cc|ccc|ccc|ccc|ccc|ccc}
    \toprule
      &  & \multicolumn{3}{c}{DUTS-TE~\cite{wang2017DUTS-TE}} & \multicolumn{3}{c}{DUT-OMRON~\cite{yang2013saliencyOMRON}} & \multicolumn{3}{c}{HKU-IS~\cite{LiYu15HKU}} & \multicolumn{3}{c}{ECSSD~\cite{shi2015hierarchicalECSSD}} & \multicolumn{3}{c}{PASCAL-S~\cite{li2014secretsPASCAL-S}}\\
    \midrule
     Method& Params. (M)& $F_{max}$ & MAE  & $E_{m}$ & $F_{max}$ & MAE & $E_{m}$ &
    $F_{max}$ & MAE &  $E_{m}$ &
    $F_{max}$ & MAE &  $E_{m}$ &
    $F_{max}$ & MAE &  $E_{m}$\\
    \midrule
    \multicolumn{17}{c}{Models with Pre-trained Backbone}\\
    \midrule
    PiCANet-R$_{\text{CVPR'18}}$~\cite{liu2018picanet}&47.22&
    0.860&0.051&0.862&
    0.803&0.065&0.841&
    0.918&0.043&0.936&
    0.935&0.046&0.913&
    0.868&0.078&0.837\\
    
    BASNet$_{\text{CVPR'19}}$~\cite{qin2019basnet}&87.06&
    0.860&0.048&0.884&
    0.805&0.056&0.869&
    0.928&0.032&0.946&
    0.942&0.037&0.921&
    0.860&0.079&0.850\\
    
    F3-Net$_{\text{AAAI'20}}$~\cite{wei2020f3net} & 26.5&
    0.891&0.035&0.902&
    0.813&0.053&0.870&
     0.937&0.028&0.953&
     0.945&0.033&0.927&
    0.882&0.064&0.863\\

    PoolNet+$_{\text{TPAMI'21}}$~\cite{Liu21PamiPoolNet} & - &
    0.889&0.037&0.896&
    0.805&0.054&0.868&
    0.936&0.030&0.953&
    0.949&0.035&0.925&
    0.892&0.067&0.859\\
    
    VST$_{\text{ICCV'21}}$~\cite{Liu_2021_ICCV_VST} & 44.48 &
    0.890&  0.037&0.892 & 
    0.825&0.058&0.861 & 
    0.942&0.029&0.953&
    0.951&0.033&0.918&
    0.890&0.062&0.846\\

    PSG$_{\text{TIP'21}}$~\cite{yang2021progressivepsg}&25.55&
    0.886&0.036&0.908&
    0.811&0.052&0.870&
    0.938&0.027&0.958&
    0.949&0.031&0.928&
    0.886&0.063&0.863\\
    EnergyTransf$_{\text{NeurIPS'21}}$~\cite{jing_ebm_sod21generative}&118.96&
    0.910&0.029&0.918&
    0.839&0.050&0.886&
    0.947&0.023&0.961&
    0.959&0.023&0.933&
    0.900&0.055&0.869\\

    RCSB$_{\text{WACV'22}}$~\cite{Ke_2022_WACV_RCSB} & 27.90& 
    0.889&0.035&0.903&
    0.810&0.045&0.856&
    0.938&0.027&0.954&
    0.944&0.033&0.923&
    0.886&0.061&0.850\\
    
    CSF-R2Net$_{\text{TPAMI'22}}$~\cite{gao2020sod100kcsf}& 36.53&
    0.890&0.037&0.897&
     0.815&0.055&0.861&
    0.935&0.030&0.952&
    0.950&0.033&0.928&
    0.886&0.069&0.855\\
    
    EDNet$_{\text{TIP'22}}$~\cite{wu2022edn} & 42.85 & 
    0.895&0.035&0.908&
    0.828&0.048&0.876&
    0.941&0.026&0.956&
    0.951&0.032&0.929&
    0.891&0.065&0.867\\
    
    PGNet$_{\text{CVPR'22}}$~\cite{xie2022pyramid} & 72.70 &
    0.917&0.027&0.922&
    0.835&0.045&0.887&
    0.948&0.024&0.961&
    0.960&0.027&0.932&
    0.904&0.056&0.878\\
    
    ICON-S$_{\text{TPAMI'22}}$~\cite{zhuge2021salienticon} & 92.40&
    0.920&0.025&0.930 & 
    0.855&0.042&0.897& 
    0.951&0.022&0.965 &
    0.961&0.023&0.932 & 
    0.906&0.051&0.875\\
    
    TRACER1$_{\text{AAAI'22}}$~\cite{lee2022tracer} & 9.96&
    0.888&  0.033&0.913 & 
    0.822&0.046&0.879 & 
    0.935&0.027&0.957 &
    0.948&0.031&0.926 & 
    0.891&0.059&0.870\\
    
    TRACER7$_{\text{AAAI'22}}$~\cite{lee2022tracer} & 66.27&
    0.927& 0.022&0.934 & 
    0.834&0.042&0.878 & 
    0.951&0.020&0.964 &
    0.959&0.026&0.927 & 
    0.911&0.049&0.880\\
    
    \midrule
    \multicolumn{17}{c}{Models without Pre-trained Backbone}\\
    \midrule
    $U^{2}$-Net$_{\text{PR'20}}$~\cite{Qin_2020_PRU2}& 44.02 &
    0.873&  0.045&0.886 & 
    0.823&  0.054&0.871&  
    0.935&0.031&0.948&
    0.951&0.033&0.924 & 
    0.868&0.078&0.845\\
    \textbf{SODAWideNet-S (Ours)} & \textbf{3.03}&
    \textbf{0.872}&\textbf{0.044}&\textbf{0.890} & 
    \textbf{0.825}&\textbf{0.054}&\textbf{0.875} & 
    \textbf{0.934}&\textbf{0.031}&\textbf{0.949} &
    \textbf{0.941}&\textbf{0.039}&\textbf{0.918} & 
    \textbf{0.868}&\textbf{0.083}&\textbf{0.849}\\
    \textbf{SODAWideNet (Ours)} & \textbf{9.03}&
    \textbf{0.883}&\textbf{0.039}&\textbf{0.895} & 
    \textbf{0.834}&\textbf{0.050}&\textbf{0.887} & 
    \textbf{0.938}&\textbf{0.028}&\textbf{0.952} &
    \textbf{0.949}&\textbf{0.037}&\textbf{0.924} & 
    \textbf{0.871}&\textbf{0.079}&\textbf{0.850}\\
    U-Net & 10.28&
    0.742&0.075&0.834 & 
    0.666&0.100&0.773 & 
    0.858&0.057&0.910 &
    0.867&0.076&0.883 & 
    0.786&0.111&0.819\\
    PGNet$^{*}$ & 72.70&
    0.823&  0.060&0.851 & 
    0.779&0.068&0.837 & 
    0.909&0.042&0.934 &
    0.916&0.054&0.907 & 
    0.839&0.094&0.824\\
    ICON-S$^{*}$ & 92.40&
    0.733& 0.080&0.818 & 
    0.704&0.082&0.811 & 
    0.837&0.071&0.894 &
    0.859&0.085&0.874 & 
    0.764&0.129&0.796\\
    TRACER-1$^{*}$ & 9.96&
    0.711&  0.092&0.801 & 
    0.704&0.087&0.800 & 
    0.833&0.069&0.890 &
    0.853&0.080&0.875 & 
    0.764&0.128&0.793\\
    \bottomrule
  \end{tabular}
  \end{adjustbox}
  \vspace{-5mm}
\end{table*}

\subsection{Datasets and Implementation Details}

We train our model on the DUTS~\cite{wang2017DUTS-TE} dataset, containing 10,553 images for training. We augment the data using horizontal and vertical flipping to obtain a training dataset of 31,659 images. We use five datasets to evaluate the proposed model. They are DUTS-Test\cite{wang2017DUTS-TE} consisting of 5019 images, DUT-OMRON\cite{yang2013saliencyOMRON} which consists of 5168 images, HKU-IS\cite{LiYu15HKU} which consists of 4447 images, ECSSD\cite{shi2015hierarchicalECSSD} which consists of 1000 images and PASCAL-S\cite{li2014secretsPASCAL-S} dataset consisting of 850 images. SODAWideNet is trained for 41 epochs on DUTS~\cite{wang2017DUTS-TE} with an initial learning rate of 0.001, and multiplied by 0.1 after 30 epochs. Two Nvidia RTX 3090 GPUs have been used to train our model with a batch size of six. Images are resized to $384 \times 384$ for training. and $416 \times 416$ for testing. We use Adam optimizer with its default parameters to update the weights. The evaluation metrics used for comparing the proposed models with prior works are the Mean Absolute Error(MAE), maximum F-measure, and the E-measure\cite{fan2018enhancedEmeasure}.

\begin{figure*}[ht!]
    \centering
    \begin{subfigure}{0.050\linewidth}
        \includegraphics[width=\textwidth]{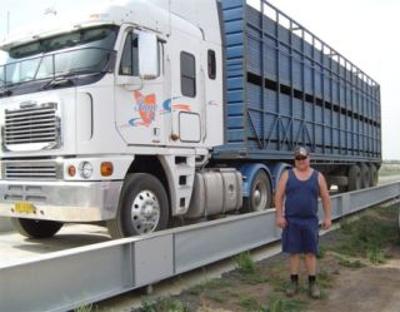}
    \end{subfigure}
    \hspace{-0.175cm}
    \begin{subfigure}{0.050\linewidth}
        \includegraphics[width=\textwidth]{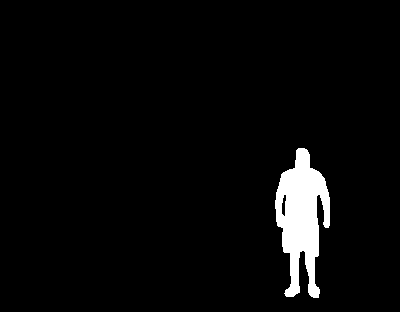}
    \end{subfigure}
    \hspace{-0.175cm}
    \begin{subfigure}{0.050\linewidth}
        \includegraphics[width=\textwidth]{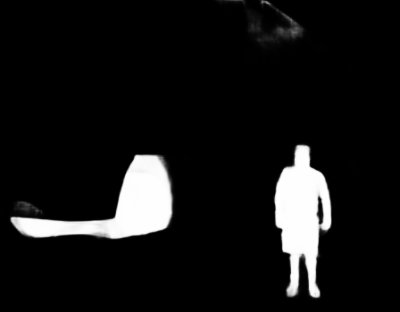}
    \end{subfigure}
    \hspace{-0.175cm}
    \begin{subfigure}{0.050\linewidth}
        \includegraphics[width=\textwidth]{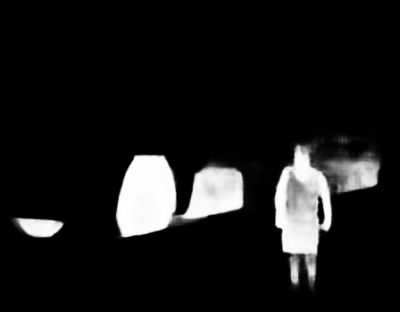}
    \end{subfigure}
    \hspace{-0.175cm}
    \begin{subfigure}{0.050\linewidth}
        \includegraphics[width=\textwidth]{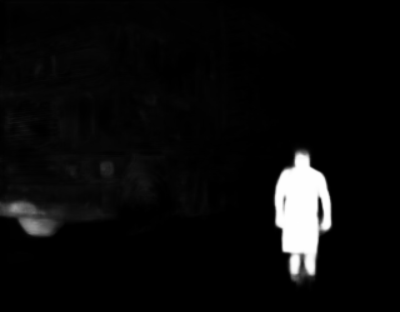}
    \end{subfigure}
    \hspace{-0.175cm}
    \begin{subfigure}{0.050\linewidth}
        \includegraphics[width=\textwidth]{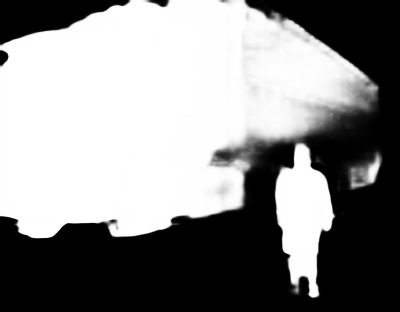}
    \end{subfigure}
    \hspace{-0.175cm}
    \begin{subfigure}{0.050\linewidth}
        \includegraphics[width=\textwidth]{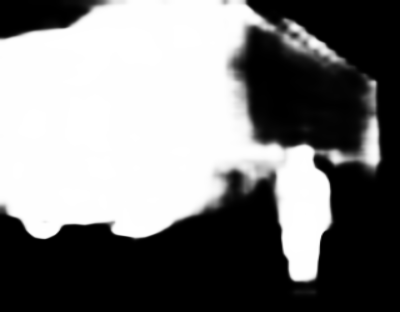}
    \end{subfigure}
    \hspace{-0.175cm}
    \begin{subfigure}{0.050\linewidth}
        \includegraphics[width=\textwidth]{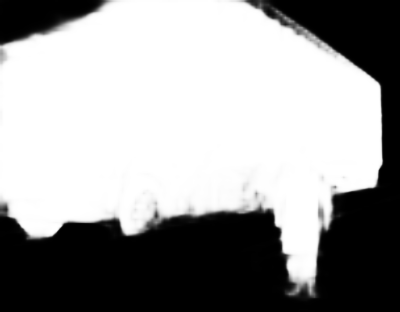}
    \end{subfigure}
    \hspace{-0.175cm}
    \begin{subfigure}{0.050\linewidth}
        \includegraphics[width=\textwidth]{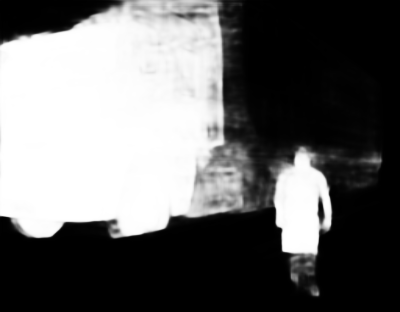}
    \end{subfigure}
    \hspace{-0.175cm}
    \begin{subfigure}{0.050\linewidth}
        \includegraphics[width=\textwidth]{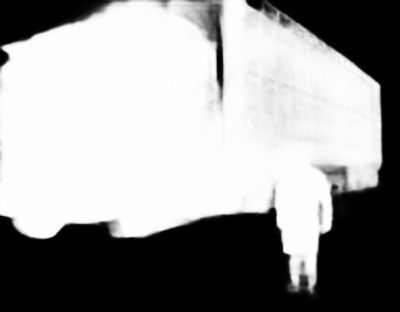}
    \end{subfigure}
    \hspace{-0.175cm}
    \begin{subfigure}{0.050\linewidth}
        \includegraphics[width=\textwidth]{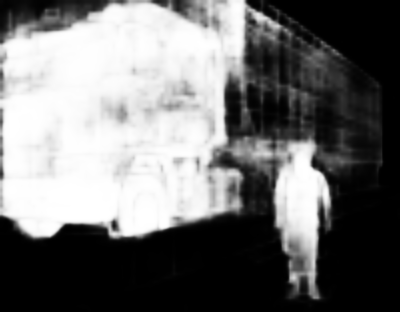}
    \end{subfigure}
    \hspace{-0.175cm}
    \begin{subfigure}{0.050\linewidth}
        \includegraphics[width=\textwidth]{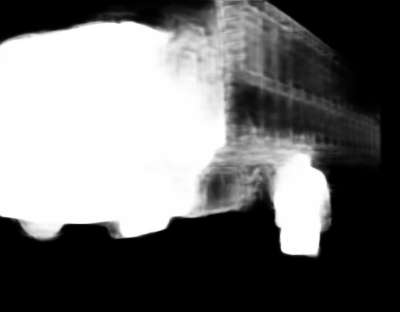}
    \end{subfigure}
    \hspace{-0.175cm}
    \begin{subfigure}{0.050\linewidth}
        \includegraphics[width=\textwidth]{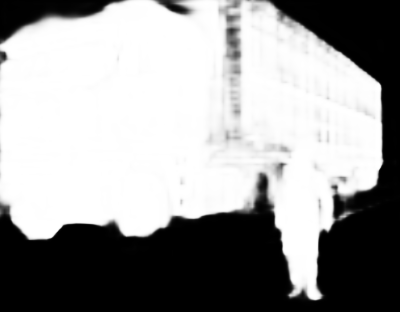}
    \end{subfigure}
    \hspace{-0.175cm}
    \begin{subfigure}{0.050\linewidth}
        \includegraphics[width=\textwidth]{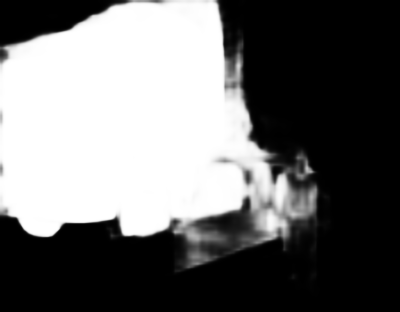}
    \end{subfigure}
    \hspace{-0.175cm}
    \begin{subfigure}{0.050\linewidth}
        \includegraphics[width=\textwidth]{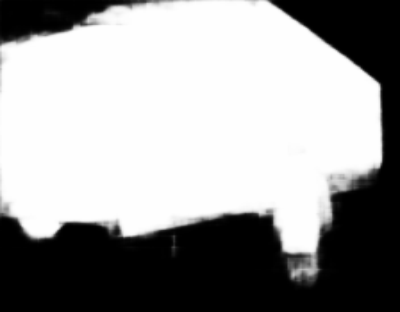}
    \end{subfigure}
    \hspace{-0.175cm}
    \begin{subfigure}{0.050\linewidth}
        \includegraphics[width=\textwidth]{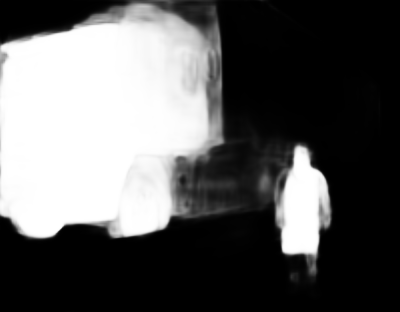}
    \end{subfigure}
    \hspace{-0.175cm}
    \begin{subfigure}{0.050\linewidth}
        \includegraphics[width=\textwidth]{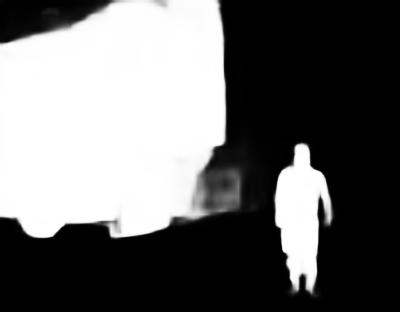}
    \end{subfigure}
    \hspace{-0.175cm}
    \begin{subfigure}{0.050\linewidth}
        \includegraphics[width=\textwidth]{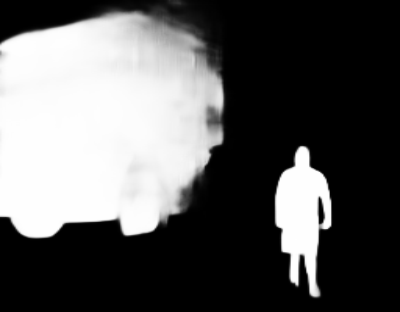}
    \end{subfigure}
    \hspace{-0.175cm}
    \begin{subfigure}{0.050\linewidth}
        \includegraphics[width=\textwidth]{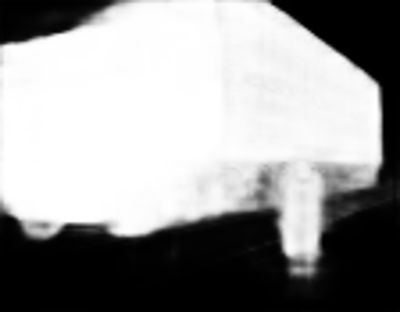}
    \end{subfigure}
    \captionsetup{labelformat=empty}
\vspace{-15mm}
\end{figure*}
\addtocounter{figure}{-1}
\begin{figure*}[h!]
    \centering
    \begin{subfigure}{0.050\linewidth}
        \includegraphics[width=\textwidth]{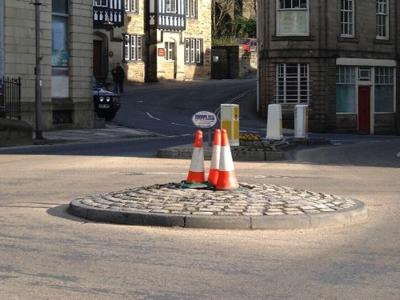}
    \end{subfigure}
    \hspace{-0.175cm}
    \begin{subfigure}{0.050\linewidth}
        \includegraphics[width=\textwidth]{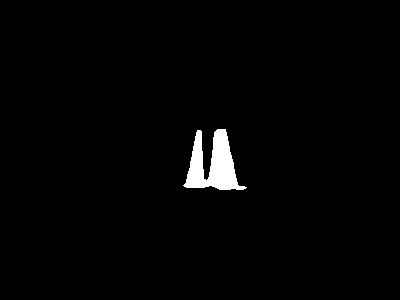}
    \end{subfigure}
    \hspace{-0.175cm}
    \begin{subfigure}{0.050\linewidth}
        \includegraphics[width=\textwidth]{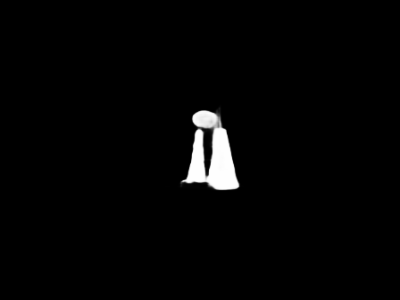}
    \end{subfigure}
    \hspace{-0.175cm}
    \begin{subfigure}{0.050\linewidth}
        \includegraphics[width=\textwidth]{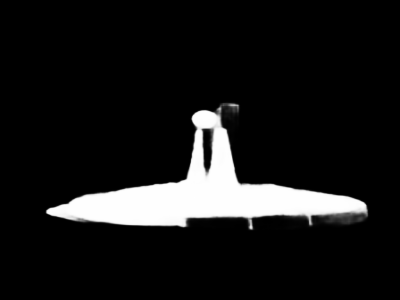}
    \end{subfigure}
    \hspace{-0.175cm}
    \begin{subfigure}{0.050\linewidth}
        \includegraphics[width=\textwidth]{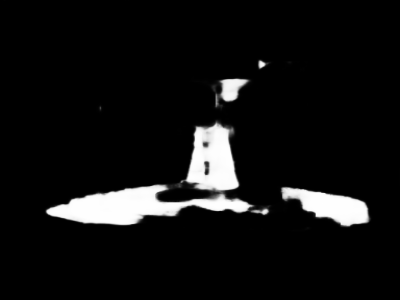}
    \end{subfigure}
    \hspace{-0.175cm}
    \begin{subfigure}{0.050\linewidth}
        \includegraphics[width=\textwidth]{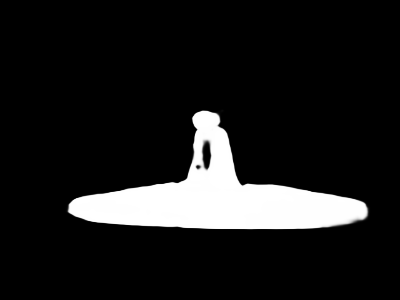}
    \end{subfigure}
    \hspace{-0.175cm}
    \begin{subfigure}{0.050\linewidth}
        \includegraphics[width=\textwidth]{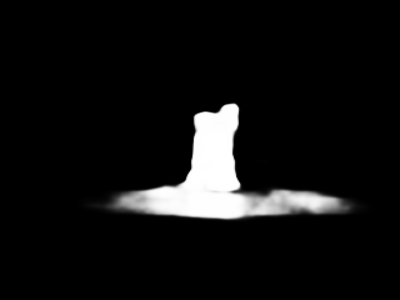}
    \end{subfigure}
    \hspace{-0.175cm}
    \begin{subfigure}{0.050\linewidth}
        \includegraphics[width=\textwidth]{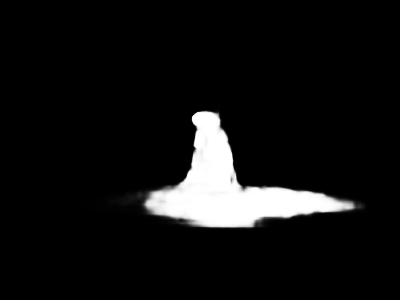}
    \end{subfigure}
    \hspace{-0.175cm}
    \begin{subfigure}{0.050\linewidth}
        \includegraphics[width=\textwidth]{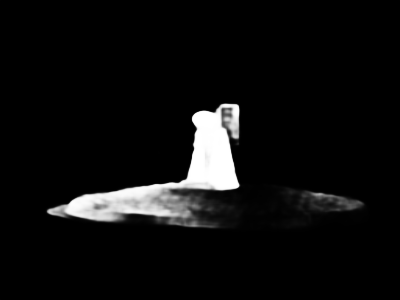}
    \end{subfigure}
    \hspace{-0.175cm}
    \begin{subfigure}{0.050\linewidth}
        \includegraphics[width=\textwidth]{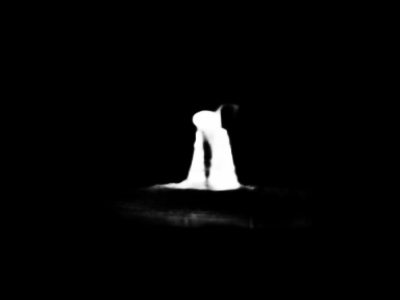}
    \end{subfigure}
    \hspace{-0.175cm}
    \begin{subfigure}{0.050\linewidth}
        \includegraphics[width=\textwidth]{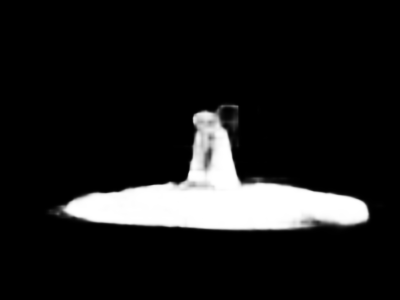}
    \end{subfigure}
    \hspace{-0.175cm}
    \begin{subfigure}{0.050\linewidth}
        \includegraphics[width=\textwidth]{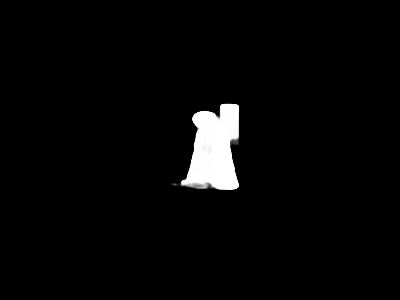}
    \end{subfigure}
    \hspace{-0.175cm}
    \begin{subfigure}{0.050\linewidth}
        \includegraphics[width=\textwidth]{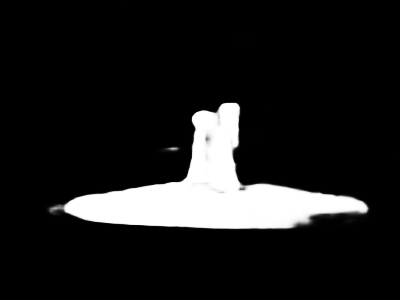}
    \end{subfigure}
    \hspace{-0.175cm}
    \begin{subfigure}{0.050\linewidth}
        \includegraphics[width=\textwidth]{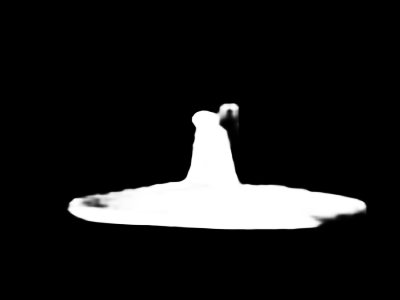}
    \end{subfigure}
    \hspace{-0.175cm}
    \begin{subfigure}{0.050\linewidth}
        \includegraphics[width=\textwidth]{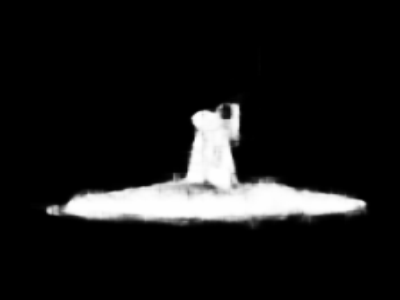}
    \end{subfigure}
    \hspace{-0.175cm}
    \begin{subfigure}{0.050\linewidth}
        \includegraphics[width=\textwidth]{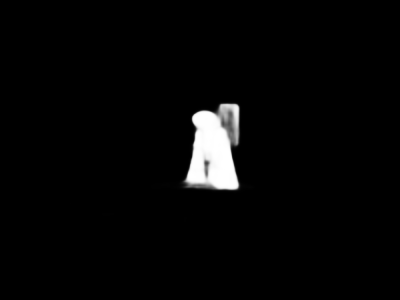}
    \end{subfigure}
    \hspace{-0.175cm}
    \begin{subfigure}{0.050\linewidth}
        \includegraphics[width=\textwidth]{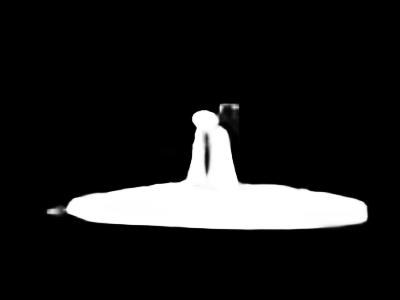}
    \end{subfigure}
    \hspace{-0.175cm}
    \begin{subfigure}{0.050\linewidth}
        \includegraphics[width=\textwidth]{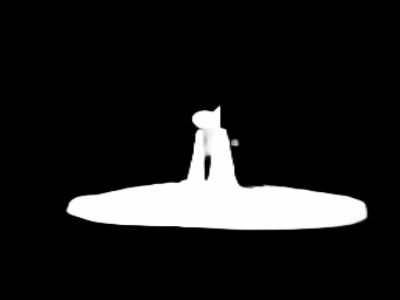}
    \end{subfigure}
    \hspace{-0.175cm}
    \begin{subfigure}{0.050\linewidth}
        \includegraphics[width=\textwidth]{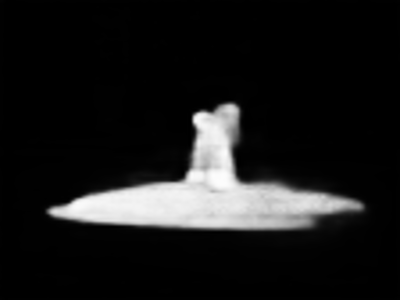}
    \end{subfigure}
\vspace{-15mm}
\end{figure*}
\addtocounter{figure}{-1}
\begin{figure*}[h!]
    \centering
    \begin{subfigure}{0.050\linewidth}
        \includegraphics[width=\textwidth]{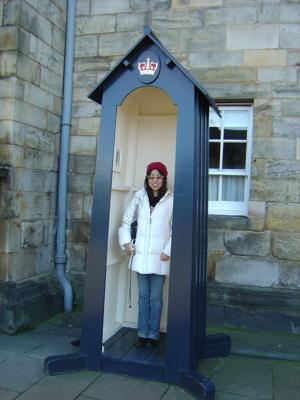}
    \end{subfigure}
    \hspace{-0.175cm}
    \begin{subfigure}{0.050\linewidth}
        \includegraphics[width=\textwidth]{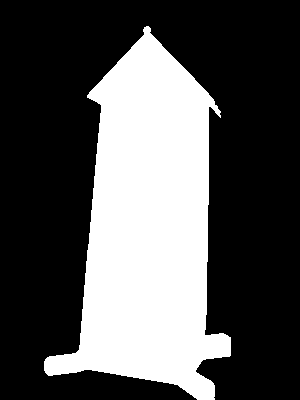}
    \end{subfigure}
    \hspace{-0.175cm}
    \begin{subfigure}{0.050\linewidth}
        \includegraphics[width=\textwidth]{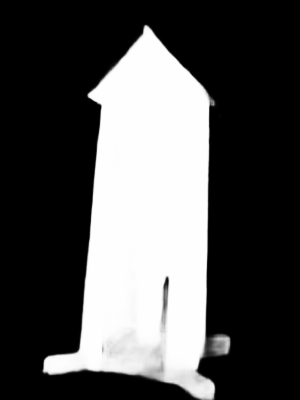}
    \end{subfigure}
    \hspace{-0.175cm}
    \begin{subfigure}{0.050\linewidth}
        \includegraphics[width=\textwidth]{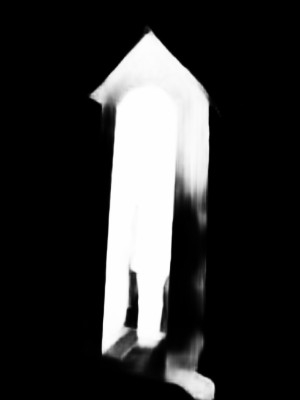}
    \end{subfigure}
    \hspace{-0.175cm}
    \begin{subfigure}{0.050\linewidth}
        \includegraphics[width=\textwidth]{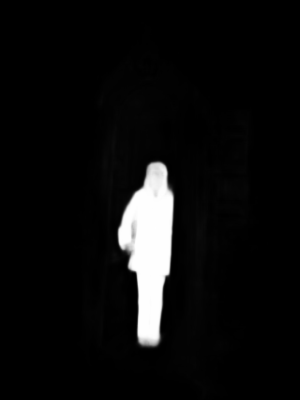}
    \end{subfigure}
    \hspace{-0.175cm}
    \begin{subfigure}{0.050\linewidth}
        \includegraphics[width=\textwidth]{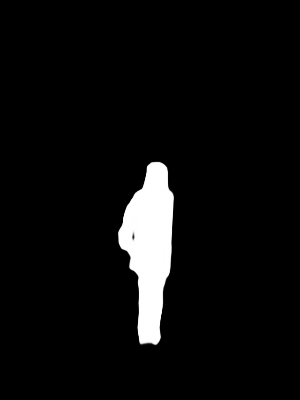}
    \end{subfigure}
    \hspace{-0.175cm}
    \begin{subfigure}{0.050\linewidth}
        \includegraphics[width=\textwidth]{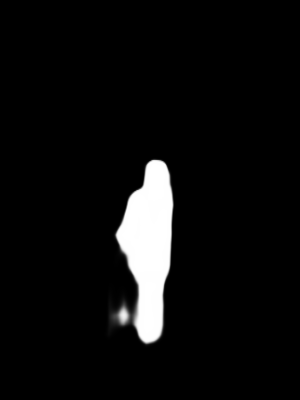}
    \end{subfigure}
    \hspace{-0.175cm}
    \begin{subfigure}{0.050\linewidth}
        \includegraphics[width=\textwidth]{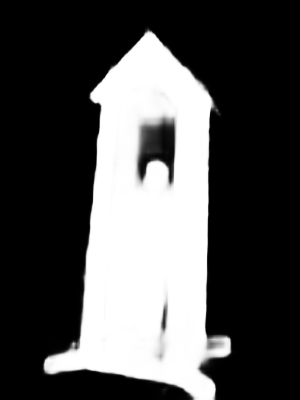}
    \end{subfigure}
    \hspace{-0.175cm}
    \begin{subfigure}{0.050\linewidth}
        \includegraphics[width=\textwidth]{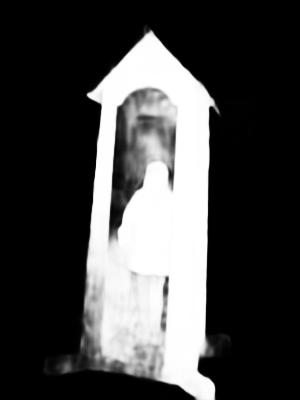}
    \end{subfigure}
    \hspace{-0.175cm}
    \begin{subfigure}{0.050\linewidth}
        \includegraphics[width=\textwidth]{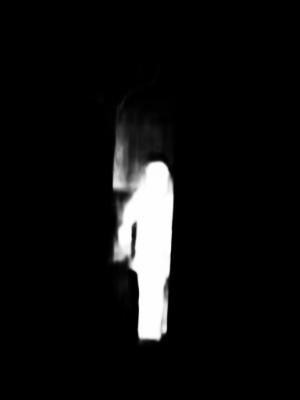}
    \end{subfigure}
    \hspace{-0.175cm}
    \begin{subfigure}{0.050\linewidth}
        \includegraphics[width=\textwidth]{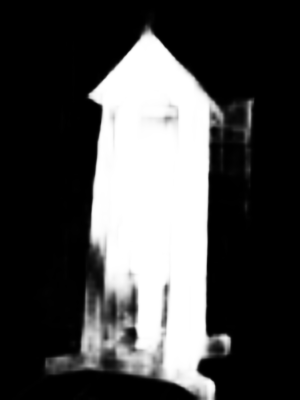}
    \end{subfigure}
    \hspace{-0.175cm}
    \begin{subfigure}{0.050\linewidth}
        \includegraphics[width=\textwidth]{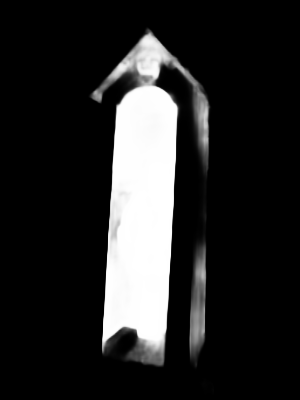}
    \end{subfigure}
    \hspace{-0.175cm}
    \begin{subfigure}{0.050\linewidth}
        \includegraphics[width=\textwidth]{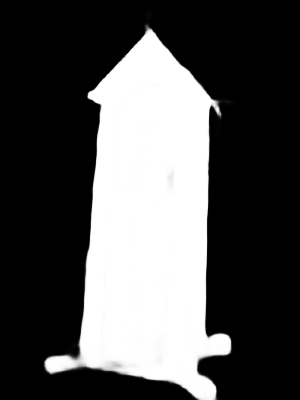}
    \end{subfigure}
    \hspace{-0.175cm}
    \begin{subfigure}{0.050\linewidth}
        \includegraphics[width=\textwidth]{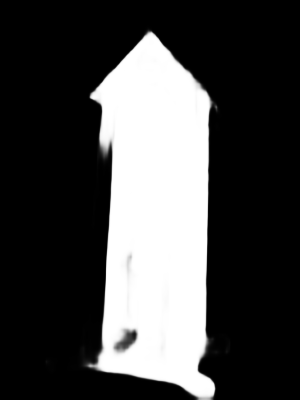}
    \end{subfigure}
    \hspace{-0.175cm}
    \begin{subfigure}{0.050\linewidth}
        \includegraphics[width=\textwidth]{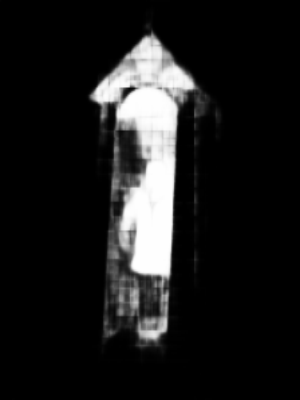}
    \end{subfigure}
    \hspace{-0.175cm}
    \begin{subfigure}{0.050\linewidth}
        \includegraphics[width=\textwidth]{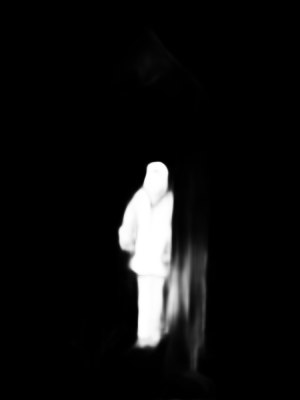}
    \end{subfigure}
    \hspace{-0.175cm}
    \begin{subfigure}{0.050\linewidth}
        \includegraphics[width=\textwidth]{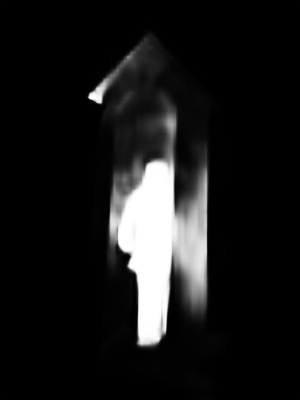}
    \end{subfigure}
    \hspace{-0.175cm}
    \begin{subfigure}{0.050\linewidth}
        \includegraphics[width=\textwidth]{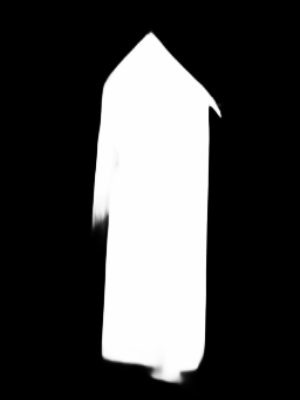}
    \end{subfigure}
    \hspace{-0.175cm}
    \begin{subfigure}{0.050\linewidth}
        \includegraphics[width=\textwidth]{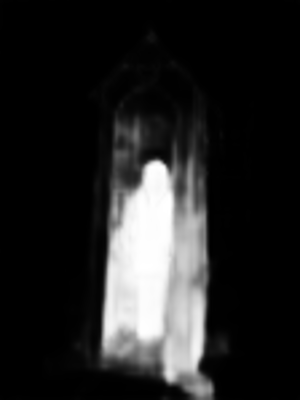}
    \end{subfigure}
    \caption{Visual comparison of 15 state-of-the-art methods.}
    \label{fig:VisualResults}
\vspace{-10mm}
\end{figure*}

\subsection{Quantitative and Visual Comparison} 
Table~\ref{tab:ResultsTable} shows the quantitative comparison against 15 state-of-the-art models for SOD. Additionally, we train PGNet, ICON, and TRACER-1 from scratch without ImageNet pre-trained weights and report their results. The proposed models perform competitively against other state-of-the-art methods. Moreover, SODAWideNet outperforms all other models that do not use a pre-trained backbone. Interestingly, the smaller SODAWideNet-S with only 3.03M, outperforms the larger $U^{2}$-Net on most metrics. On the other hand, models relying on features from ImageNet pre-trained backbones suffer catastrophically without pre-trained weights. For example, the performance gap between using a pre-trained backbone and training from scratch ranges from 10\% for PGNet to 18.87\% for ICON on the DUTS dataset. On the other datasets, this performance gap tends to reduce to 13\%. Finally, U-Net outperforms ICON and TRACER-1 on most metrics, clearly showing the challenges of training from scratch and the robustness of encoder-decoder-style models for dense prediction tasks. 

Figure~\ref{fig:VisualResults} shows some representative examples of our model predictions compared to other SOTA models. The first two images are the input and the ground truth respectively. The images from the third to last column follow the same order as in Table~\ref{tab:ResultsTable} from SODAWideNet to PiCANet-R. SODAWideNet performs well on smaller objects in a challenging environment (rows 1 and 2), segmenting large objects (rows 3). Additionally, the SODAWideNet-S performs comparatively well against other state-of-the-art models.

\subsection{Influence of MSA, MRFFAM, and LPM}

Table~\ref{tab:components} illustrates the influence of each component in our model architecture. It shows that contour supervision as an auxiliary task significantly improves model performance. Furthermore, this additional supervision also enables adding more complexity to the proposed model. MSA is the next component to affect the model performance profoundly. Without MSA, the $F_{max}$ score drops by 1.2\%, showing the significance of using attention from the initial layers. Furthermore, removing MRFFAM in the hybrid and conv block also had a substantial impact (1.0\% and 0.5\%, respectively), indicating the effectiveness of long-range convolutional features. * indicates the removal of MRFFAM in the conv block. Finally, the model also suffers significantly (-0.6\%) without LPM, showing the importance of local features.
\begin{table}[h!]
\centering
  \begin{tabular}{|c|c|c|c|c|c|}
    \hline
      Contours  & MSA& MRFFAM & LPM & $F_{max}$ & MAE \\  
    \hline
     $\times$& \checkmark & \checkmark & \checkmark& 0.868 & 0.045 \\
    \checkmark  & $\times$ & \checkmark & \checkmark&  0.871 & 0.044 \\
    \checkmark  & \checkmark & $\times$& \checkmark&  0.873 & 0.043\\
    \checkmark  & \checkmark & \checkmark& $\times$&  0.877 & 0.042\\
    \checkmark & \checkmark & $\times$*& \checkmark&  0.878 & 0.041\\
    \checkmark &  \checkmark & \checkmark& \checkmark& \textbf{0.883} & \textbf{0.039}\\
    \hline
  \end{tabular}
  \caption{Influence of individual components in SODAWideNet.}
  \label{tab:components}
  \vspace{-13mm}
 \end{table}

\section{Conclusion}

We propose a novel encoder-decoder model for Salient Object Detection using dilated convolutions and self-attention without ImageNet pre-training. Inspired by Vision transformers, we use large convolution kernels at every layer to obtain semantic information from farther regions. This strategy contrasts modern convolutional backbones like ResNet-50, which use small convolution kernels with a deep network. Furthermore, to induce self-attention into our network through Multi-Scale Attention (MSA) that computes attention at higher resolutions. Finally, the competitive results with a parameter-efficient model reveal a promising direction toward designing robust vision models without expensive ImageNet pre-training.
 
%
%
\bibliographystyle{splncs04}
\bibliography{mybib}
%




\end{document}